\documentclass[manuscript,screen]{acmart}

\usepackage{subcaption}

\usepackage{xspace}
\usepackage{booktabs}
\usepackage{makecell}
\usepackage{xcolor, colortbl}
\usepackage{tabularx}
\usepackage{hyperref}

\newcommand{\eg}{\emph{e.g.,}\xspace}
\newcommand{\etal}{\emph{et al.}\xspace}
\newcommand{\ie}{\emph{i.e.,}\xspace}
\definecolor{Gray}{gray}{0.925}

\DeclareMathOperator*{\argmin}{arg\,min}

\usepackage{listings}
\usepackage{xcolor}

\colorlet{punct}{red!60!black}
\definecolor{background}{HTML}{EEEEEE}
\definecolor{delim}{RGB}{20,105,176}
\colorlet{numb}{magenta!60!black}

\lstdefinelanguage{json}{
    basicstyle=\normalfont\ttfamily,
    numbers=left,
    numberstyle=\scriptsize,
    stepnumber=1,
    numbersep=8pt,
    showstringspaces=false,
    breaklines=true,
    frame=lines,
    backgroundcolor=\color{background},
    literate=
     *{0}{{{\color{numb}0}}}{1}
      {1}{{{\color{numb}1}}}{1}
      {2}{{{\color{numb}2}}}{1}
      {3}{{{\color{numb}3}}}{1}
      {4}{{{\color{numb}4}}}{1}
      {5}{{{\color{numb}5}}}{1}
      {6}{{{\color{numb}6}}}{1}
      {7}{{{\color{numb}7}}}{1}
      {8}{{{\color{numb}8}}}{1}
      {9}{{{\color{numb}9}}}{1}
      {:}{{{\color{punct}{:}}}}{1}
      {,}{{{\color{punct}{,}}}}{1}
      {\{}{{{\color{delim}{\{}}}}{1}
      {\}}{{{\color{delim}{\}}}}}{1}
      {[}{{{\color{delim}{[}}}}{1}
      {]}{{{\color{delim}{]}}}}{1},
}

\AtBeginDocument{%
  \providecommand\BibTeX{{%
    \normalfont B\kern-0.5em{\scshape i\kern-0.25em b}\kern-0.8em\TeX}}}

\setcopyright{acmcopyright}
\copyrightyear{2022}
\acmYear{2022}
\acmDOI{XXXXXXX.XXXXXXX}

\begin{document}

\title{Self-Contained Entity Discovery from Captioned Videos}

\author{Melika Ayoughi}
\email{m.ayoughi@uva.nl}
\orcid{??}
\affiliation{%
  \institution{University of Amsterdam}
  \city{Amsterdam}
  \country{The Netherlands}
}

\author{Pascal Mettes}
\affiliation{%
  \institution{University of Amsterdam}
  \city{Amsterdam}
  \country{The Netherlands}
}
\email{p.s.m.mettes@uva.nl}

\author{Paul Groth}
\affiliation{%
  \institution{University of Amsterdam}
  \city{Amsterdam}
  \country{The Netherlands}
}
\email{p.t.groth@uva.nl}

\renewcommand{\shortauthors}{Ayoughi, et al.}

\begin{teaserfigure}
  \includegraphics[width=\textwidth]{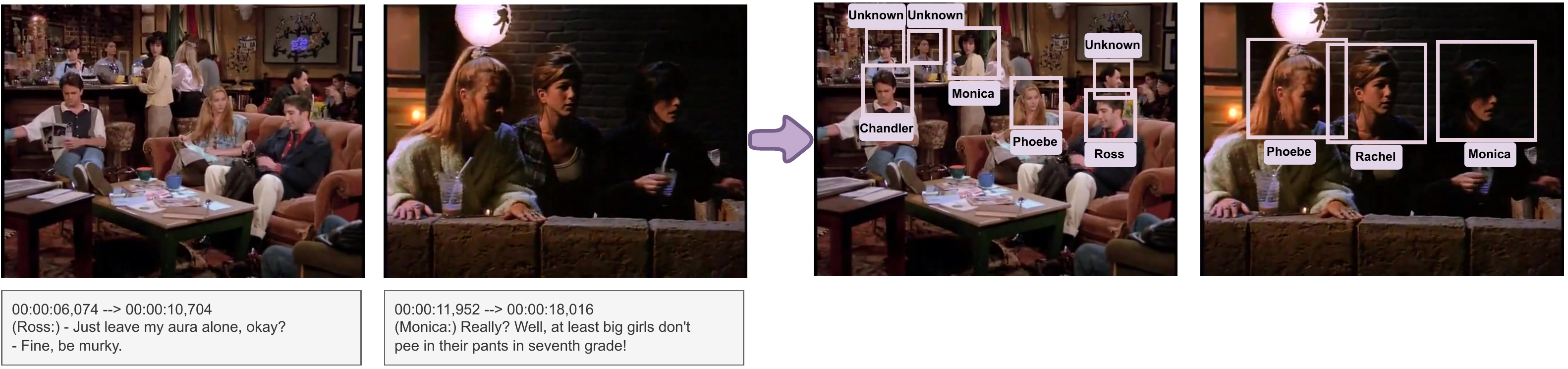}
  \caption{Is it possible to uncover the visual entities occurring in a collection of captioned videos without requiring task-specific supervision or external knowledge sources? We introduce self-contained video entity discovery, where we seek to localize and align visual entities with named entities solely from corresponding captions. Left: video frames and corresponding captions given as model input. Right: localized visual entities with automatically assigned named entities as model output.}
  \label{fig:problem_definition}
\end{teaserfigure}

\begin{abstract}
This paper introduces the task of visual named entity discovery in videos without the need for task-specific supervision or task-specific external knowledge sources. Assigning specific names to entities (e.g. faces, scenes, or objects) in video frames is a long-standing challenge. Commonly, this problem is addressed as a supervised learning objective by manually annotating faces with entity labels. To bypass the annotation burden of this setup, several works have investigated the problem by utilizing external knowledge sources such as movie databases. While effective, such approaches do not work when task-specific knowledge sources are not provided and can only be applied to movies and TV series. In this work, we take the problem a step further and propose to discover entities in videos from videos and corresponding captions or subtitles. We introduce a three-stage method where we (i) create bipartite entity-name graphs from frame-caption pairs, (ii) find visual entity agreements, and (iii) refine the entity assignment through entity-level prototype construction. To tackle this new problem, we outline two new benchmarks \textsc{SC-Friends}  and \textsc{SC-BBT} based on the Friends and Big Bang Theory TV series. Experiments on the benchmarks demonstrate the ability of our approach to discover which named entity belongs to which face or scene, with an accuracy close to a supervised oracle, just from the multimodal information present in videos. Additionally, our qualitative examples show the potential challenges of self-contained discovery of any visual entity for future work. The code and the data are available on GitHub.\footnote{\url{https://github.com/Melika-Ayoughi/Self-Contained-Video-Entity-Discovery}}
\end{abstract}





\maketitle

\section{Introduction}
Vision and language play an important role for learning from videos, whether it be classifying labels from a fixed word list~\cite{li2018unified, feichtenhofer2019slowfast, feichtenhofer2016spatiotemporal,tran2018closer} or generating video descriptions~\cite{chen2019generating, bin2021multi,yan2019stat, rohrbach2015dataset,tapaswi2015book2movie,torabi2015using}. A recurring limitation is that such tasks require manual annotations. As a promising new direction, several recent works have proposed to learn video representations by aligning visual and textual embeddings, \eg by projecting both to a shared embedding space~\cite{huang2018learning, sun2019videobert,mithun2018learning} and by aligning video fragments with textual narrations~\cite{alayrac2016unsupervised, tan2021look, miech2020end, miech2019howto100m,huang2020caption,6156449, Everingham2006HelloMN}. Here, we take inspiration from this research direction and seek to discover which entities from captions and scripts belong to which visual entities in video frames, see Figure~\ref{fig:problem_definition}.

To recognize and predict labels for visual entities in videos, two directions are dominant. The first direction follows a supervised perspective, where labels are assigned to individual entities (e.g. faces) in training videos~\cite{guo2016ms, kemelmacher2016megaface, cao2018vggface2}. A model is in turn trained on the annotated training data and applied in test videos. While effective, such a setup requires manual supervision for each new entity. To make entity prediction more scalable, the second direction seeks to exploit task-specific knowledge sources to provide additional information, focusing predominantly on person entities. Commonly, movie databases are used where main characters with images of corresponding actors are crawled to obtain labelled data to train a model on~\cite{huang2020movienet, guo2018curriculumnet, 6909814, 7780475, 45493, 46299, zhuang2017attend, jiang2018mentornet, 8604143, huang2020caption}. This alleviates the need for manual annotations for each entity, but its scope is limited to popular movies and TV series, for which supporting information is available. In this work, we investigate whether it is possible to uncover the names of any visual entity - from persons to scenes - using only videos, captions, and generic pre-trained visual recognition models. Here, captions can come from scripts, subtitles, or speech-to-text. The ability to discover entities in videos in a self-contained manner not only alleviates the need for task-specific knowledge or annotations but also allows for applicability to general videos.

A core challenge is that the captions and associated video frames in the general are only weakly correlated, making a direct matching between the two impossible. To that end, we outline a three-stage method for self-contained video entity discovery. In the first stage, we create bipartite entity-name graphs from frame-caption pairs. In the second stage, we discover visual entity agreements through over-clustering and use the bipartite graphs to assign entity names to clusters. In the third stage, we alleviate the bias towards highly frequent named entities in the cluster assignments by constructing entity prototypes and reassigning cluster entities with respect to the prototypes. Our approach is general and can be applied to any type of entity.

To evaluate our approach, we furthermore introduce two new benchmarks for self-contained entity discovery from captioned videos: \textsc{SC-Friends} and \text{SC-BBT} (self-contained Friends and self-contained Big Bang Theory). Experiments on the new benchmarks show that our approach is able to discover people entities and the name that belongs to them with an accuracy close to a supervised oracle. Importantly, this is done without the need for any manual annotations or use of external knowledge sources, while also outperforming approaches from related tasks such as the MediaEval challenge \cite{bredin2016multimodal}. We run additional quantitative and qualitative analyses to show the effect of the over-clustering and the size of the entity bounding boxes for local entities such as persons. The ability to automatically link named entities to visual entities from captioned videos alone has multiple real-world applications, including the annotation of large video archives for free \cite{brown2021automated} and entity instance retrieval in e-commerce \cite{zhan2021product1m} and robotics \cite{grauman2022ego4d} settings.


Summarized, the contributions of this work are as follows:
\begin{itemize}
\item We introduce the task of self-contained video entity discovery, where named entities need to be uncovered and assigned to visual entities using only a collection of captioned videos;

\item We introduce two new benchmarks, \textsc{SC-Friends} and \text{SC-BBT}, to evaluate performance on the self-contained video entity discovery task;

\item We propose a three-stage approach for self-contained video entity discovery that automatically filters, assigns, and debiases the match between named and visual entities;

\item We perform comparative evaluations and ablation studies to highlight the potential of our approach over current approaches.


\end{itemize}



\section{Related Work}

\subsection{Multimodal video understanding}
Video understanding constitutes a long-standing challenge in multimedia and computer vision, covering a wide range of research tasks, including but not limited to action recognition and localization~\cite{feichtenhofer2019slowfast, feichtenhofer2016spatiotemporal,tran2018closer}, video description \cite{rohrbach2015dataset,tapaswi2015book2movie,torabi2015using,zhu2015aligning}, and video question answering \cite{chen2019generating, bin2021multi,yan2019stat,jang2017tgifqa,kim2017deepstory,lei2019tvqa,mun2017marioqa}. In recent years, a popular direction in video understanding is to learn representations by exploiting the multimodal nature of videos, specifically visual and textual information \cite{sun2019videobert}. For example, Bain \etal \cite{bain2020condensed} and Zellers \etal \cite{zellers2019recognition} propose joint vision-language benchmarks for text-to-video retrieval and question answering. Similarly, Mahon \etal \cite{mahon2020knowledge} propose to generate knowledge graphs of video clips based on their descriptions, while Miech \etal~\cite{miech2019howto100m} learn representation by aligning millions of videos and captions. We take inspiration from these advances and seek to discover entities for faces by exploiting the multimodal nature of captioned videos, rather than relying on manual supervision.

Besides a multimodal focus, a growing body of literature also seeks to go beyond short videos and standard action labels.
Long-form video understanding \cite{wu2021longform} exemplifies this and aims at understanding the full picture of a long-form video, including the storyline of a movie, the relationships among the characters, the message conveyed by their creators, and the aesthetic styles.
For understanding beyond standard labels, several papers have investigated the automatic learning and decomposition of tasks from instructional videos \cite{alayrac2016unsupervised, tan2021look, miech2020end}. Other work seeks to uncover the scene graph and the interactions of a video, where nodes denote objects and edges denote activities \cite{zellers2018neural,yang2018graph, woo2018linknet, li2017scene, tian2020part, ji2021detecting, kukleva2020learning}. We also aim to learn from long videos beyond standard labels. However, where most work focuses on the activities and interactions of generic instance types (\eg person), we aim to find the specific named entity of each instance detected in videos (\eg Ross).


\subsection{Entity recognition in movies and television}
Movies and TV series have proven to be a fruitful source of information for video understanding \cite{huang2020movienet,vicol2018moviegraphs}. Specifically, movies and TV series provide a valuable testbed for discovering entities in videos. A wide range of datasets taken from movies have been introduced, such as
MovieQA, \cite{tapaswi2016movieqa}, the Large Scale Movie Description Challenge \cite{rohrbach2017movie},  Hollywood2 \cite{marszalek09}, and AVA \cite{gu2018ava}. MovieNet \cite{huang2020movienet} contains movies with their trailers, photos, plot descriptions, characters, scene boundaries, description sentences, place and action, and cinematic style tags. However, the movies are not publicly available due to copyright reasons.

A reoccurring goal in video understanding is recognizing entities and the relations between entities in the form of graphs.
The nodes in such graphs denote entities of different types such as people, objects, and locations, while the edges denote relations such as relationships and activities. MovieGraphs \cite{vicol2018moviegraphs} provides, graph-based annotations of social situations in movie clips, consisting of people, their emotional and physical attributes, their relationships, and the interactions between them. The High-Level Video Understanding challenge \cite{curtis2020hlvu} is a dataset of open source movies and their high-level knowledge graph, plus queries that need reasoning and retrieving non-visual concepts. The key and starting point of these graphs is given by the person entities detected in video frames.

Identifying characters and named entities in movies and TV series has been tackled from multiple directions.
Commonly, face recognition models are trained in a supervised manner on large-scale crowd-sourced annotated datasets \cite{guo2016ms, kemelmacher2016megaface, cao2018vggface2}, where the specific entity is given for each training face. Given a trained model, the entities of new faces can then be directly inferred. While effective, a supervised setup is manually expensive and needs to be redone for every new entity. Therefore, several works have investigated entity discovery in faces from videos through entity-specific knowledge sources. These knowledge sources are movie databases (e.g. IMDb) with general knowledge about movie and TV series including the key characters and the corresponding actors. By crawling such databases, it is possible to obtain visual training examples of entities for training~\cite{guo2018curriculumnet, 6909814, 7780475, 45493, 46299, zhuang2017attend, jiang2018mentornet}. Haq \etal \cite{8604143} for example exploit IMDb to detect starring characters in movies, while Huang \etal~\cite{huang2020caption} utilize tagged images from the web. Brown \etal~\cite{brown2021automated} combine image-search engines and IMDb name lists as evidence sources. The use of entity-specific knowledge sources alleviates the need for manual supervision of entities. 

The 2015 MediaEval challenge on person discovery in broadcast news \cite{poignant2015multimodal, bredin2016multimodal} attempts to eliminate the use of external knowledge sources for multimedia indexing. The goal of the challenge is to find the names of people who can be simultaneously seen and heard within a shot where the list of people is not known a priori and must be discovered from media content using text overlay or speech transcripts. Our goal of self-contained entity discovery in videos differs from MediaEval in three ways: (i) we seek to assign a name to each individual visual entity, rather than provide a list of names per shot; (ii) we do not assume that an entity is simultaneously seen and mentioned; (iii) our approach is not specific to persons can be applied to other entity types such as objects and scenes. To that end, we provide two new benchmarks and we show that approaches tailored to MediaEval do not generalize to our challenging setting.

Our work also relates to research in face detection in movies and TV series. Earlier work in this field focus on clustering and improving face representations. Kalogeiton \etal introduce must-link and cannot-link constraints to the hierarchical clustering method \cite{kalogeiton2020constrained}.
Other work proposes to leverage labels obtained from clustering along with video constraints to learn discriminative face features \cite{sharma2020clustering}. Tapaswi \etal~\cite{tapaswi2019video} focus on clustering with unknown number of clusters to include minor characters. Our work differs in that we want to uncover the particular entity associated with each face.

\section{Method}
\begin{figure*}
    \centering
    \includegraphics[width=\linewidth]{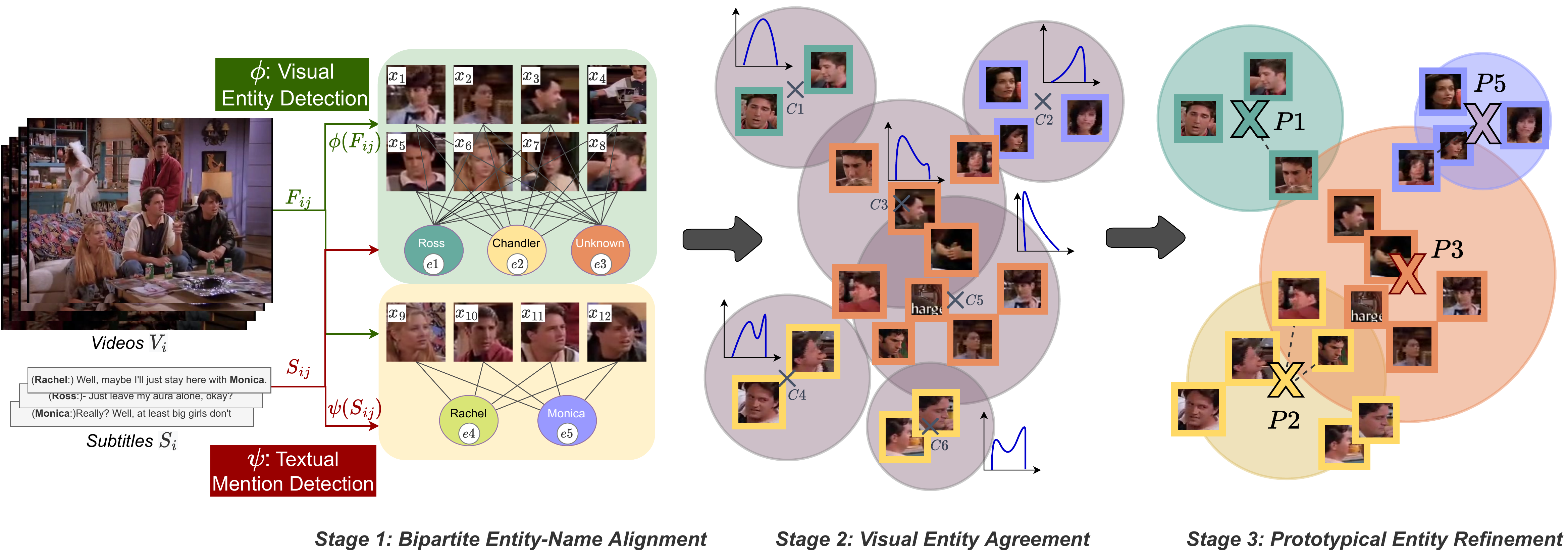}
    \caption{Method for self-contained video entity discovery: In the first stage, bipartite graphs are built by densely matching visual box entities extracted from video frames to named entities from a mention detection model. The second stage performs over-clustering, and for each cluster, the most-occurring entity label is aggregated. In the last stage, samples from the \textit{most frequent} entity label are reassigned based on their distance to class prototypes (Here, the orange entity label). }
    \label{fig:method}
\end{figure*}


For the task of self-contained video entity discovery, we are given a collection of $N$ videos $\mathcal{V} = \{V_1,..,V_N\}$, where each video $V_i$ consists of a set of video frames and a set of textual descriptions (e.g subtitles), with the goal to uncover the specific names of entities that occur in $\mathcal{V}$.
We propose a three-stage approach to tackle this problem, see Figure \ref{fig:method}:
\begin{itemize}
\item \emph{Stage 1: Bipartite entity-name alignment.} 
First, we process both modalities to obtain named entities from textual descriptions and visual entities from corresponding frames. For each description-frame pair, a set of visual entity and named entity nodes are created and densely connected to form a collection of bipartite graphs, one graph for each description-frame pair, which provides initial cues about the entity names that correspond to boxes. However, this alignment is non-unique, incomplete, and noisy. The labels (i.e. names) assigned to each box after this stage are called {\em weak labels}.
\item \emph{Stage 2: Visual entity agreement.} Second, we seek to improve the matching between visual entities from frames and named entities from textual descriptions by taking advantage of the similarities between visual entities. This is done by over-clustering the visual embeddings and then using the alignments from the bipartite graphs of the previous stage to aggregate clusters into entity distributions. The most frequently occurring entity name is then selected as the new name for each box this is termed the {\em cleansed label}.
\item \emph{Stage 3: Prototypical entity refinement.} By aggregating the most-occurring entity name for each cluster, the most frequent entity name will be over-represented in the entity distribution. This aggregation biases the entity discovery towards most-frequent labels. As a third stage we address this bias by computing a per-entity prototype, followed by a refinement of the entity assignment for boxes of the most frequent prototype based on the minimal prototypical distance.


\end{itemize}

The proposed approach is generic and applies to any type of entity. For the rest of this section, we use person entities as running example to elaborate the three stages.

\subsection{Stage 1: Bipartite entity-name alignment}
For a given video $V_i$, the corresponding input is given by a set of frames $F_i = \{F_{ij}\}_{j=1}^{T_i}$ with $T_i$ the number of frames for video $i$ and a set of textual descriptions $S_i = \{S_{ij}\}_{j=1}^{T_i}$.
In the first stage, faces are extracted from all frames using a pre-trained generic face detection network $\phi(\cdot)$~\cite{facenet}. From the textual descriptions, entities are extracted using a pre-trained generic named entity recognition network $\psi(\cdot)$~\cite{dbmdzhuggingface}.\footnote{Here, we use an NER model because we focus on person entities. Generally, we can use any model that identifies mentions of entities in text.} While each detected face becomes a separate instance, only unique named entities from the text are considered. Let the set of unique entities for video $i$ be denoted as:
\begin{equation}
E = \bigcup_{i=1}^{N} \bigcup_{j=1}^{T_i} \psi(S_{ij}).
\end{equation}
Notably, our method groups non-recurring minor characters, such as side-characters in busy street scenes, in an \emph{unknown} category, which can be done automatically through a cutoff on the distribution frequency in $E$. Then the method densely connects the set of detected faces in each frame with the detected unique entities in the corresponding textual description in a bipartite manner. Let $G_{ij}$ denote the bipartite graph for frame $j$, defined as:
\begin{equation}
G_{ij} = (\phi(F_{ij}), E[\psi(S_{ij})], \epsilon_{ij}),
\end{equation}
with
\begin{equation}
\phi(F_{ij}) = \{\mathbf{x}_l\}_{l=1}^{D_{ij}}, \quad \mathbf{x}_l \in \mathbb{R}^{d}
\end{equation}
where $D_{ij}$ denotes the number of detected faces in frame $j$ of video $i$ and with $d$ the feature dimensionality in face embedding space. Here, $E[\cdot]$ denotes the set indexing of the entities and $\epsilon_{ij}$ denotes the dense connectivity from all faces to entities in the frame.

The output of the first stage is given as:
\begin{equation}
G = \{G_{ij} \ | \ \forall_{i \in N}, \forall_{j \in T_i} \}.
\end{equation}
$G$ denotes a set of bipartite graphs between faces and entities over all videos and intuitively provides the initial links between faces and entities, namely where they occur across modalities at the same time. This output is however limited in multiple aspects: (i) each face can be connected to multiple entities if they appear in the same textual description; (ii) the entity matching is incomplete as a person can be present in a scene and not mentioned by name; and (iii) the entity matching is noisy, as a person not present in the scene can be mentioned by name.

\subsection{Stage 2: Visual entity agreement}
Since the initial face-entity alignment is simultaneously overcomplete, incomplete, and noisy, it can not be directly used for self-contained discovery. The alignment does provide initial weak cues about which face might belong to which entities. But how can we steer this alignment without explicit supervision? Our solution lies in the faces and their feature embeddings. Following the first stage, each detected face becomes a unique instance. Key person entities will however re-occur throughout an entire video collection. While it is impractical to directly infer the entity of a face from the initial alignment, if there is some signal in $G$, then faces that are close together in the face embedding space tend to have similar matching entities.

Based on this inter-face entity agreement assumption, faces are aggregated using their feature embedding through over-clustering. Let $\mathcal{C} = \{C_1, C_2, ..., C_k\}$ with $k > |E|$ denote a set of cluster centroids, optimized as:
\begin{equation}
\mathcal{C} = \argmin_{C^{\star}} \sum_{i,j,l=1,1,1}^{N,T_i,D_{ij}} \sum_{c \in C^{\star}} || \phi(F_{ij})_l - c ||^2,
\end{equation}
which is obtained through a clustering algorithm applied over the set of faces from all videos. With this setup, one can determine the entity of a cluster as the one which is matched most frequently to the faces in the cluster based on $G$. The discovered entity for each face is then determined by the entity of the cluster it belongs to. With this stage, we utilize the ``wisdom of the crowd" in face embedding space combined with the weak cues from $G$ to obtain a unique and robust named entity for each individual face.



\subsection{Stage 3: Prototypical entity refinement}

An important component in the entity discovery of the second stage is to use more clusters than unique entities, which empirically works better as it is able to handle pockets of similar faces which can form due to physical transformations of characters over time.  Here, the \emph{most frequent} class in each cluster will represent the whole cluster after the assignment. Thus, the most frequent prototype will be over-represented which will change the prototype frequency ratios to one overly-represented prototype and other under-represented prototypes. As a consequence, the output of the second stage is sensitive to entity imbalance. To avoid a bias towards the \emph{most frequent} entity, we refine the entity assignment using notions from Prototypical Networks~\cite{snell2017prototypical}. For entity $e$, let $F_e$ denote the set of all faces whose predicted entity from the second stage is equal to $e$. We then define its prototype as:
\begin{equation}
P_e = \frac{1}{|F_e|} \sum_{\mathbf{x} \in F_e} \mathbf{x}.
\end{equation}



After obtaining the prototypes for all entities, each face $f$ with embedding $\mathbf{x}_f$ whose initial predicted entity was given as the \emph{most frequent} entity is refined as:
\begin{equation}
E^*(f) = \argmin_{e \in E} ||\mathbf{x}_f - P_e||^2.
\end{equation}
In other words, after the second stage we have a high precision over all entities, but a low recall since many faces are assigned to the \emph{most frequent} category. Using the prototypical refinement of the third stage, those faces that were assigned to the \emph{most frequent} but are far away from its prototype will be reassigned, providing a better balance between precision and recall and improving the performance for under-represented categories. 




\section{Self-Contained Video Entity Discovery Benchmark (SC-VED)}

To evaluate and test our method, we construct two datasets: \textsc{SC-Friends} (self-contained Friends) and \textsc{SC-BBT} (self-contained Big Bang Theory) from the video frames and subtitles of the TVQA datasets \cite{lei2019tvqa,lei2020tvqa+}. 
TVQA is a dataset tailored to the video question answering task, which
publicly provides the 3fps video frames of $5$ complete long-form TV series
.
For quantitative evaluations, we focus on person entities and corresponding faces in video frames due to the availability of generic models for face recognition and person named entity detection. We also include qualitative evaluations on scene entities in the paper.

\subsection{\textsc{SC-Friends}} 
We select the first $5$ episodes of season $1$ of the \textit{Friends} TV series from TVQA. This consists of $1598$ frames and corresponding subtitles. We discard frame-subtitle pairs where there are either no faces in the frame or no people present in the subtitle.
Subsequently, we create ground-truth annotations by detecting all the faces within the dataset using a pre-trained face recognition model. Using the frequency distribution of person entities in the benchmark, we automatically create seven classes: Joey, Rachel, Ross, Phoebe, Monica, Chandler, and an unknown category. The unknown category combines all named entities in the long-tail of the entire entity set from stage 1. We manually annotate each face with a corresponding label from these classes. 
Note: the annotations are only for evaluation and ignored in our approach. 
The resulting set of annotations contains $3386$ unique face bounding boxes.

\subsection{\textsc{SC-BBT}}
For the second benchmark, we leverage existing annotations in the TVQA+ dataset \cite{lei2020tvqa+}. TVQA+ is an extension to the TVQA dataset that also includes character bounding boxes of the \textit{Big Bang Theory} series. It provides video frames of all episodes  up to season $10$ episode $13$ (the series ran for $12$ seasons). In the train and validation set of the TVQA+, characters and objects in the scene related to the question-answering task are annotated with a bounding box around them and a ground-truth label. 
We preprocess the TVQA+ annotations in 4 steps.
\begin{enumerate}
\label{text:preprocess}
    \item For each annotated bounding box in TVQA+, the bounding box is removed from ground-truth annotations if there are no matching subtitles. Every subtitle has a start and finish time; if the frame is not between beginning and end, it does not have a matching subtitle.
    \item There are $3015$ ground-truth labels including $2413$ objects and $602$ people. We manually group them into objects and people categories. Bounding boxes with labels categorized as objects are discarded.
    \item The labels of the TVQA+ dataset have dictation errors and include many variations of character names for instance, Sheldon is written in $27$ different ways, including Sheldons, sHeldon, Shedlon, Heldon, etc. We perform fuzzy matching between any pair of main character entity names and other labels. For each character entity, we keep labels with a fuzzy ratio higher than $70$. We then manually check those, discard them if needed, and use the fuzzy matched labels for relabelling. In the last step, if a ground-truth label is not among the fuzzily matched labels, it is labeled as \emph{unknown}.
\end{enumerate}



After performing these preprocessing steps, we arrive at $122264$ ground truth face bounding box annotations from $76116$ number of video frames. Each bounding box is labeled with either one of the $8$ main characters or \emph{unknown}. The main characters (again determined automatically based on frequency of occurrence as named entities) are Penny, Leonard, Howard, Stuart, Sheldon, Bernadette, Amy, and Raj, leaving the rest as a separate unknown category. 


\section{Experimental Setup}

\subsection{Baselines}
To compare the performance of our method, we employ four different baselines.
\begin{enumerate}
    \item \emph{Oracle:} This is the supervised upper bound of the self-contained entity discovery task. Here, the classifier layer of the pretrained face recognition network is trained in a supervised manner with ground truth bounding boxes and labels. For SC-BBT this oracle is tuned to focus on the detected face instead of the full-body bounding box given by the TVQA+ dataset, as this provides far superior performance.
    In this case, if no face is detected in the bigger box annotation, the original bounding box is used, and if multiple faces are detected, one face is chosen randomly.
    \item \emph{Bipartite Experiment:} Here, the outputs of the first stage of our method are used to train a  multi-label classifier.
    \item \emph{LIMSI:} This is the unsupervised baseline inspired by the clustering-based method proposed in \cite{poignant2015limsi} for the MediaEval challenge \cite{le2017towards} with minor changes to adapt to our problem. Here, we apply the first stage of our method and in the second stage a direct matching policy \cite{poignant2012unsupervised} from the LIMSI solution. In the direct matching policy, first, faces with exactly one co-occurring name are directly matched. Then, the visual entity agreement is performed as per our method on the remaining unnamed faces.




\end{enumerate}


\subsection{Details for our method}
\subsubsection{Evaluation of Different Stages}
\label{sec:evaluation_of_different_stages}
We compare different stages of our method. We note that because our method is unsupervised, $100\%$ of the data is employed. For \textsc{SC-Friends} this means episodes $1$ to $5$ of season $1$, and for \textsc{SC-BBT} it includes all seasons $1$ to $10$.
To predict an entity label per face after stage 1, we randomly select one of the matched named entities in the corresponding bipartite graph. If no named entity is extracted, then it is labeled as \emph{unknown}. This intermediate result tells us whether there is any signal in the initial bipartite matching at all. For the other stages, we can directly use the assigned entity label as prediction for each face.

\subsubsection{Implementation Details}
\label{sec:implementation_details}
In the first stage of our method, each frame goes through a pretrained Inception Resnet (V1) model \cite{facenet} that extracts faces of $160 \times 160$ pixels. Simultaneously, the corresponding subtitle goes through a pretrained named entity recognition transformer model \cite{dbmdzhuggingface}. This model outputs $4$ types of named entities, including Person, Organization, Location, and Miscellaneous. We filter the person named entities from all named entities. If a person entity is split into multiple tokens, we concantenate them together in a single string (i.e. mention). Each face is then matched with all the person named entities from its corresponding subtitle. 
For the \textsc{SC-Friends} experiments, the Agglomerative Clustering method with Euclidean distance metric and Ward linkage is used with $20$ clusters in the second stage. For the \textsc{SC-BBT} experiments, because of the high number of annotations, the Agglomerative Clustering is infeasible due to memory issues; thus, k-means with $30$ clusters is used.

\subsection{Scene entity discovery} \label{sec:scene_entity_discovery} The task of video entity discovery is a generic task and can include different kinds of entities. To showcase that our approach is not limited to person entities, we perform an extra experiment with scenes as entities.
For this setting, the mention detector \cite{sanh2019hierarchical} extracts potential named entities from the subtitles. The SEER model \cite{goyal2022vision} pretrained on places dataset is given the complete frame to extract scene features. The second and third stage of the method are applied just like before. Since the textual mention detector model extracts many mentions that are not named place entities, we filter the textual mentions based on their frequencies and being a specific scene location in the given TV series. We group the mentions that refer to the same scene entity into $13$ scene entity names including: the kitchen (Monica's kitchen), the hallway, the hospital, the living room (Monica's living room), central perk, the airport, London, Barbados, the museum, the gym, the beach, the beach house, and the balcony (Monica's balcony).
The full taxonomy grouping is given in the Github repository. 
Since the frequency of scene entity names is low, in the first stage, we match each entity name with the subsequent $4$ frames except for the unknown entity names that are matched with only $1$ frame.

\subsection{Evaluation metrics, environment and splits}
We report the following metrics to evaluate different stages of our method and baselines. The per-class accuracy calculates the accuracy for each character separately, the macro accuracy takes the mean of per-class accuracies. The micro accuracy calculates the sample accuracy irrespective of which class they belong to. This metric could show high results in cases where the most-frequent class, for instance, \emph{unknown} performs well and others performs poorly. We also report the macro average of precision, recall, and F1 scores.
When evaluating on the \textsc{SC-BBT} dataset, we take the bounding box with the highest intersection over union (IoU) with the ground-truth bounding box in TVQA+ to compare labels.

We train the supervised baselines (oracle, tuned oracle and bipartite) with $80\%$ of the data and the unsupervised baseline (LIMSI) with $100\%$ of the data. We test both on $20\%$ of the data. In \textsc{SC-Friends}, we train on episodes $1$ to $4$ and test on episode $5$ in season $1$. In \textsc{SC-BBT}, we train on seasons $1$ to $8$ and test on seasons $9$ and $10$. To make the results of our method and the baselines comparable, we optimize our method with $100\%$ of the data and evaluate on the same $20\%$ as the baselines,\ie for \textsc{SC-Friends} episodes $1$ to $5$ in season $1$ for optimization and episode $5$ for evaluation. For \textsc{SC-BBT} seasons $1$ to $10$ for optimization and seasons $9$ and $10$ for evaluation.

Experiments were performed using one GeForce 1080Ti, 11GB GDDR5X. For all supervised experiments, we train the classifier layer and fix the other layers for $100$ epochs and a batch size of $256$. We initialize the learning rate to $0.0012$ with a decay rate of $0.1$ and weight decay of $0.0001$ at epoch $95$. Every face input is normalized. The default settings are used for clustering and the face embeddings are extracted from the layer before the classifier.

\section{Results}

\begin{table}[t]
\centering
\caption{Evaluating and comparing the three stages of our approach  on \textsc{SC-Friends}. Our approach outperforms baselines that use the bipartite labels from stage 1 to train a multi-label classifier and baselines from the MediaEval challenge The final performance after 3 stages is close to the oracle upper bound, highlighting the effectiveness of our approach.}
\label{tab:friends_accuracy}
\begin{tabular}{lccccc}
\toprule
Approach & \multicolumn{2}{c}{Accuracy} & Precision & Recall & F1\\
& Micro & Macro & & &\\
\midrule
\rowcolor{Gray}
\textbf{Baselines} & & & & &\\
Bipartite labels & 0.38 & 0.15 & 0.34 & 0.15 & 0.10\\
LIMSI \cite{poignant2015limsi} & 0.50 & 0.46 & 0.48 & 0.46 & 0.46 \\
Oracle labels & 0.80 & 0.82 & 0.79 & 0.82 & 0.80\\
\midrule
\rowcolor{Gray}
\textbf{This paper} & & & & &\\
Stage 1 & 0.35 & 0.25 & 0.30 & 0.25 & 0.26\\
Stage 1+2 & 0.70 & 0.69 & \textbf{0.86} & 0.69 & 0.75\\
Stage 1+2+3 & \textbf{0.73} & \textbf{0.76} & 0.76 & \textbf{0.76} & \textbf{0.75}\\
\bottomrule
\end{tabular}
\end{table}


\begin{table}[t]
\centering
\caption{Evaluating and comparing the three stages of our approach on \textsc{SC-BBT}. Akin to the results on the \textsc{SC-Friends} benchmark, we outperform competitive baselines, making a step towards oracle performance without the need for any supervision.}
\label{tab:bbt_accuracy}
\begin{tabular}{lccccc}
\toprule
Approach & \multicolumn{2}{c}{Accuracy} & Precision & Recall & F1\\
& Micro & Macro & & &\\
\midrule
\rowcolor{Gray}
\textbf{Baselines} & & & & &\\
Bipartite labels & 0.14 & 0.11 & 0.23 & 0.11 & 0.03\\
LIMSI \cite{poignant2015limsi} & 0.53 & 0.51 & 0.53 & 0.51 & 0.50 \\
Oracle labels & 0.74 & 0.72 & 0.76 & 0.72 & 0.74\\
\midrule
\rowcolor{Gray}
\textbf{This paper} & & & & &\\
Stage 1 & 0.38 & 0.36 & 0.43 & 0.36 & 0.37\\
Stage 1+2 & \textbf{0.59} & 0.56 & \textbf{0.67} & 0.57 & \textbf{0.59}\\
Stage 1+2+3 & 0.56 & \textbf{0.59} & 0.53 & \textbf{0.59} & 0.54\\
\bottomrule
\end{tabular}
\end{table}

The overall results of the experiments on the \textsc{SC-Friends} and \textsc{SC-BBT} datasets are shown in Table \ref{tab:friends_accuracy} and \ref{tab:bbt_accuracy} respectively. On the \textsc{SC-Friends} dataset our method achieves a $76\%$ macro accuracy which compares favorably with the Oracle performance of 82\%, which is a fully supervised upper bound. 
Just using weak labels from stage 1 for training in a supervised manner (the Bipartite Experiment baseline) does not achieve good results. This is likely due to the high noisy ratio in the weak labels and because of the high ratio of \emph{unknown} labels. Thus, the model learns to predict every face as \emph{unknown} and results in an accuracy of $100\%$ for \emph{unknown} and near $0\%$ for other classes. 

Comparing our results to the LIMSI baseline, both on the \textsc{SC-Friends} and the \textsc{SC-BBT} our method outperforms the LIMSI baseline with 30\% and 8\% respectively on recall. In \textsc{SC-BBT} the difference is smaller (59\% vs. 51\%), likely due to better co-occurrence of character names and their faces. In LIMSI, a strong assumption is made about the co-occurrence of a specific name and a face. This holds for less noisy captions with perfect frame matchings however it does not hold for all applications like in news data. The per-class accuracies are constantly better compared to LIMSI for both datasets with the exception of Sheldon, a consequence of the third stage of our method where the most-frequent label can be re-assigned to other prototypes. We conclude that our approach provides high accuracy, precision, and recall for entity discovery in videos without the need for any supervision or external knowledge source.

\subsection{Contribution of each stage}
Each stage of our method is key for improving performance. This is particularly evident in the performance on the \textsc{SC-Friends} dataset. The Stage 1 macro accuracy on this dataset is $11\%$ points higher than random labeling given $7$ labels (i.e $14\%$ accuracy). The improvement of stage 1 over random labeling signals that there is indeed a weak cue in the bipartite face-entity matching. The improvement after Stage 2 is substantial, reaching the accuracy of $69\%$, since here we take advantage of the multimodal nature of videos and leverage both similarity in the visual and textual cues. With Stage 2, we optimise for precision, and with Stage 3, we optimize for recall. Thus, there is a $10$ point drop in precision but a corresponding $10$ point increase in recall. The F1 score remains the same. In addition, the macro accuracy increases by $7$ percentage points because the per-class accuracies improve for every class except for \emph{unknown}. In comparison, the micro accuracy increases only $3$ points, since the \emph{unknowns} comprise a significant proportion of samples.  

For the \textsc{SC-BBT}, again, both Stage 1 and Stage 2 are critical for improvements in performance. Achieving an $409\%$ performance improvement over random after both stages.
The prototypical entity refinement (Stage 3) optimizes for recall, thus achieving lower results for precision, micro accuracy and F1 score because the precision of a large proportion of data is affected. This stage is not as effective in this dataset but also does not harm the results. This is probably due to the higher ratios of both \emph{Sheldon} and \emph{unknown} in comparison to \textsc{SC-Friends} where \emph{unknown} is the dominating label. This is because the ground-truth annotations in \textsc{SC-BBT} are more centered around the main characters that occur in the question-answer pairs, while in \textsc{SC-Friends}, the face recognition model extracts many false-positive faces.

\subsection{Per-class accuracy}
The results of per-class accuracy are shown in Table \ref{tab:friends_per_class} and \ref{tab:bbt_per_class} respectively. Again, we can see that every stage of the method improves per-class accuracies for \textsc{SC-Friends}, while in the last stage since we are reassigning \emph{unknowns} to other entities and optimizing recall for entities other than \emph{unknown}, the accuracy of \emph{unknowns} drops from $84\%$ to $63\%$. For \textsc{SC-BBT}, the same constant improvement is seen over all classes except for the \emph{Sheldon} class that drops from $97\%$ to $69\%$, because here the \emph{Sheldon} entities are reassigned to other entity prototypes.

Over-represented characters such as Ross, Rachel and Howard perform well in both datasets. However, Penny who is one of the main characters in the Big Bang Theory series completely disappears from the entities after the second stage. Likely, this problem comes from the named entity recognition in the first stage. Since the name Penny is also a unit of currency, the NER model only detects a low proportion of Penny as a person named entity. The frequency of this entity is $0.05\%$ of the most-frequent entity (Sheldon) after the bipartite alignment, while it should be almost the same frequency. Thus, Penny is never among the most-frequent entities in the second stage so it naturally disappears from the entities and results in poor accuracy.

\begin{table}[]
\begingroup 
\setlength{\tabcolsep}{3pt}
\centering
\caption{\textsc{SC-Friends} accuracy per class for our approach and the baselines. Balancing the entities in stage 3 improves the accuracy for the main characters, at the cost of a lower accuracy for the other named entities grouped into the unknown category.}
\label{tab:friends_per_class}
\begin{tabular}{lccccccc}
\toprule
 & \multicolumn{7}{c}{\textbf{Per-class accuracy}}\\
 &  Joey & Rachel &   Ross &   Phoebe &  Monica &  Chandler &   Unknown\\
\midrule
\rowcolor{Gray}
\textbf{Baselines} & & & & & & &\\
Bipartite labels & 0.00 & 0.00   & 0.05 & 0.00   & 0.02   & 0.00     & 1.00\\
LIMSI \cite{poignant2015limsi} & 0.43 & 0.36 & 0.60 & 0.45 & 0.47 & 0.30 & 0.59\\
Oracle labels & 0.78 & 0.82   & 0.92 & 0.84   & 0.86   & 0.80     & 0.72\\
\hline
\rowcolor{Gray}
\textbf{This paper} & & & & & & & \\
Stage 1  & 0.17 & 0.17   & 0.27 & 0.14   & 0.20   & 0.19     & 0.62\\
Stage 1+2            & 0.54 & 0.67   & 0.82 & 0.71   & 0.64   & 0.63     & \textbf{0.84}\\
Stage 1+2+3           & \textbf{0.71} & \textbf{0.83}   & \textbf{0.91} & \textbf{0.80}   & \textbf{0.73}   & \textbf{0.70}     & 0.63\\
\bottomrule
\end{tabular}
\endgroup
\end{table}


\begin{table}[]
\begingroup 
\setlength{\tabcolsep}{3pt}
\caption{\textsc{SC-BBT} accuracy per class or our approach and the baselines. In this dataset, Sheldon is the most frequent entity and rebalancing this entity benefits all other entities. The exception here is Penny, which fails to achieve any results in unsupervised settings since the name is not picked up as a named entity.}
\label{tab:bbt_per_class}
\begin{tabular}[t]{lccccccccc}
\toprule
 & \multicolumn{9}{c}{\textbf{Per-class accuracy}}\\
 & Penny & Leonard & Howard & Stuart &  Sheldon & Bernadette & Amy &  Raj &  Unknown\\
 \midrule
\rowcolor{Gray}
\textbf{Baselines} & & & & & & & & &\\
Bipartite labels & 0.00 & 0.00   & 0.00 & 0.00   & 0.00   & 0.00 & 0.00 & 0.00    & 0.99 \\
LIMSI \cite{poignant2015limsi} & 0.01 & 0.58 & 0.62 & 0.63 & 0.82 & 0.55 & 0.55 & 0.53 & 0.32\\
Oracle labels & 0.82   &  0.78  &  0.75  &   0.60  &  0.85  & 0.75    &  0.72   &  0.77   &  0.47 \\
\hline
\rowcolor{Gray}
\textbf{This paper} & & & & & & & & &\\
Stage 1 &   0.01    &      0.37   &    0.40    &   0.45     &    0.49     &      0.36      &  0.31   &  0.30   &  0.59\\
Stage 1+2     &   0.00    &    0.60     &    0.69    &   0.65     &  \textbf{0.97}       & 0.64  & 0.64    &   0.58  &   0.32\\
Stage 1+2+3    &  0.00     &     \textbf{0.66}   &    \textbf{0.74}   &    \textbf{0.67}    &  0.69       &      \textbf{0.73}      &  \textbf{0.70}   &   \textbf{0.72}  &     \textbf{0.37}\\
\bottomrule
\end{tabular}
\endgroup
\end{table}

\begin{table*}[]
\begingroup 
\setlength{\tabcolsep}{2.5pt}
\caption{Impact of the size of face bounding boxes on performance (stage 2) on SC-Friends. Enlarging the scope of the face boxes to include some context around the face improves entity accuracy, precision, and recall.}
\label{tab:ablation}
\begin{tabular}{lcccccccccccc}
\toprule
 & \multicolumn{7}{c}{\textbf{Per-class accuracy}} & \multicolumn{2}{c}{\textbf{Accuracy}} & \textbf{Precision} & \textbf{Recall} & \textbf{F1}\\
                       & Joey & Rachel & Ross & Phoebe & Monica & Chandler & Unk. & Micro & Macro & & & \\ \midrule
Tight boxes & 0.33   & 0.47     & 0.83   & 0.47     & 0.54     & 0.42       & 0.87      & 0.66                    & 0.56   &    0.81  & 0.56 & 0.62        \\
\rowcolor{Gray}
Contextual boxes  & 0.54   & 0.67     & 0.82   & 0.71     & 0.64     & 0.63       & 0.84      & 0.74                    & 0.69   &  0.86  & 0.69 & 0.75 \\ \bottomrule
\end{tabular}
\endgroup
\end{table*}

\subsection{Ablation studies}
Below, we analyze other key parameters that impact the performance of our method.

\subsubsection{Size of face bounding boxes} Here, we investigated the impact of the size of the detected faces. We consider the size of the bounding box around the face as an important factor since bigger bounding boxes include more contextualized information such as hair and parts of clothing, while the tight boxes include only the face components, which makes it harder to detect individuals. 

To test this, we compare the contextual boxes used in our experiments with tight bounding boxes which are $20$ pixels smaller from each side. Table \ref{tab:ablation} depicts the results of the first $2$ stages of our method on tight and contextual bounding boxes for \textsc{SC-Friends} dataset. All results are higher for the contextual boxes except for the \emph{unknown} class.

\begin{figure}
\centering
\begin{subfigure}[t]{.4\linewidth}
  \includegraphics[width=\linewidth]{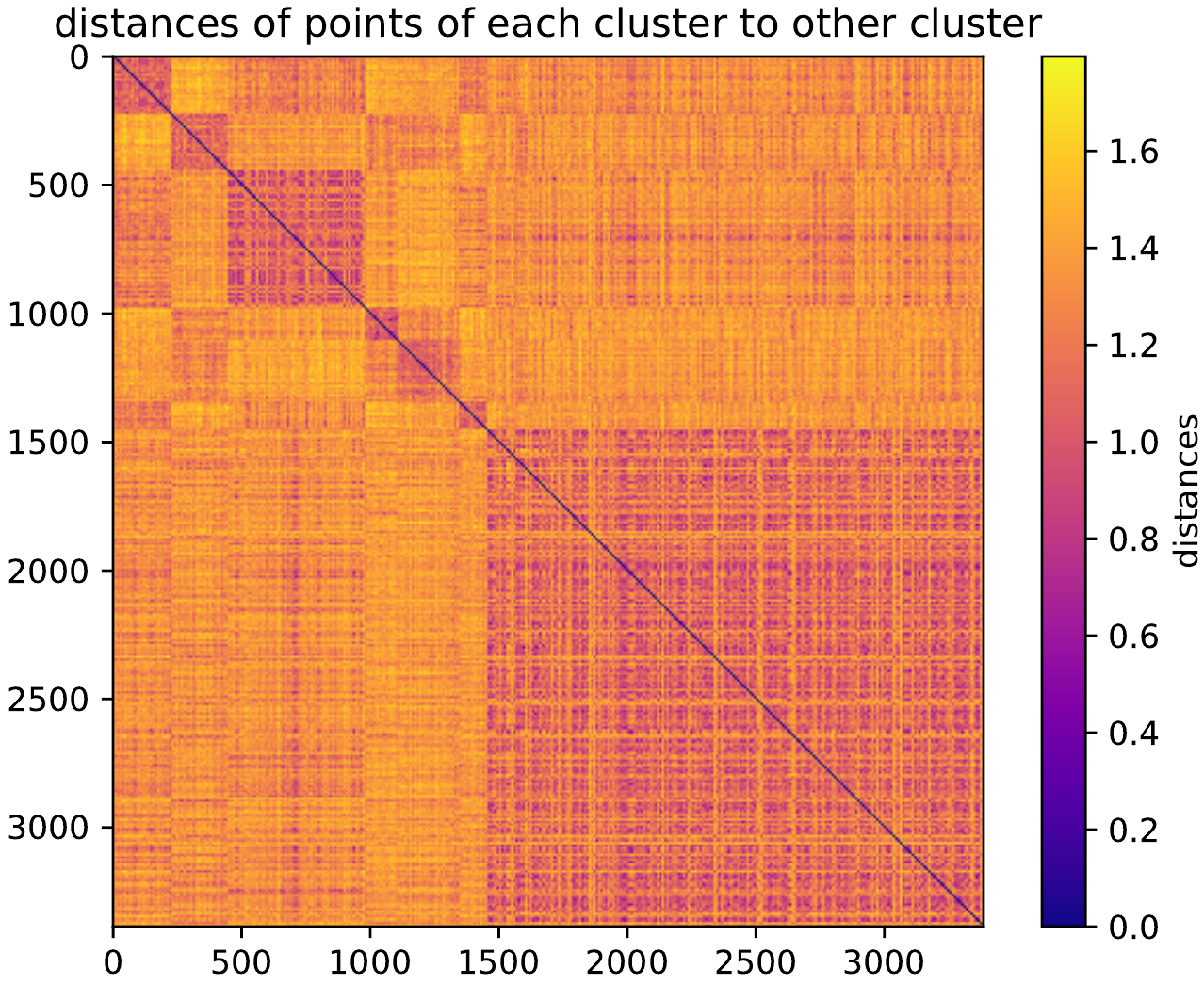}
  \caption{Tight}
\end{subfigure}
\begin{subfigure}[t]{.4\linewidth}
  \includegraphics[width=\linewidth]{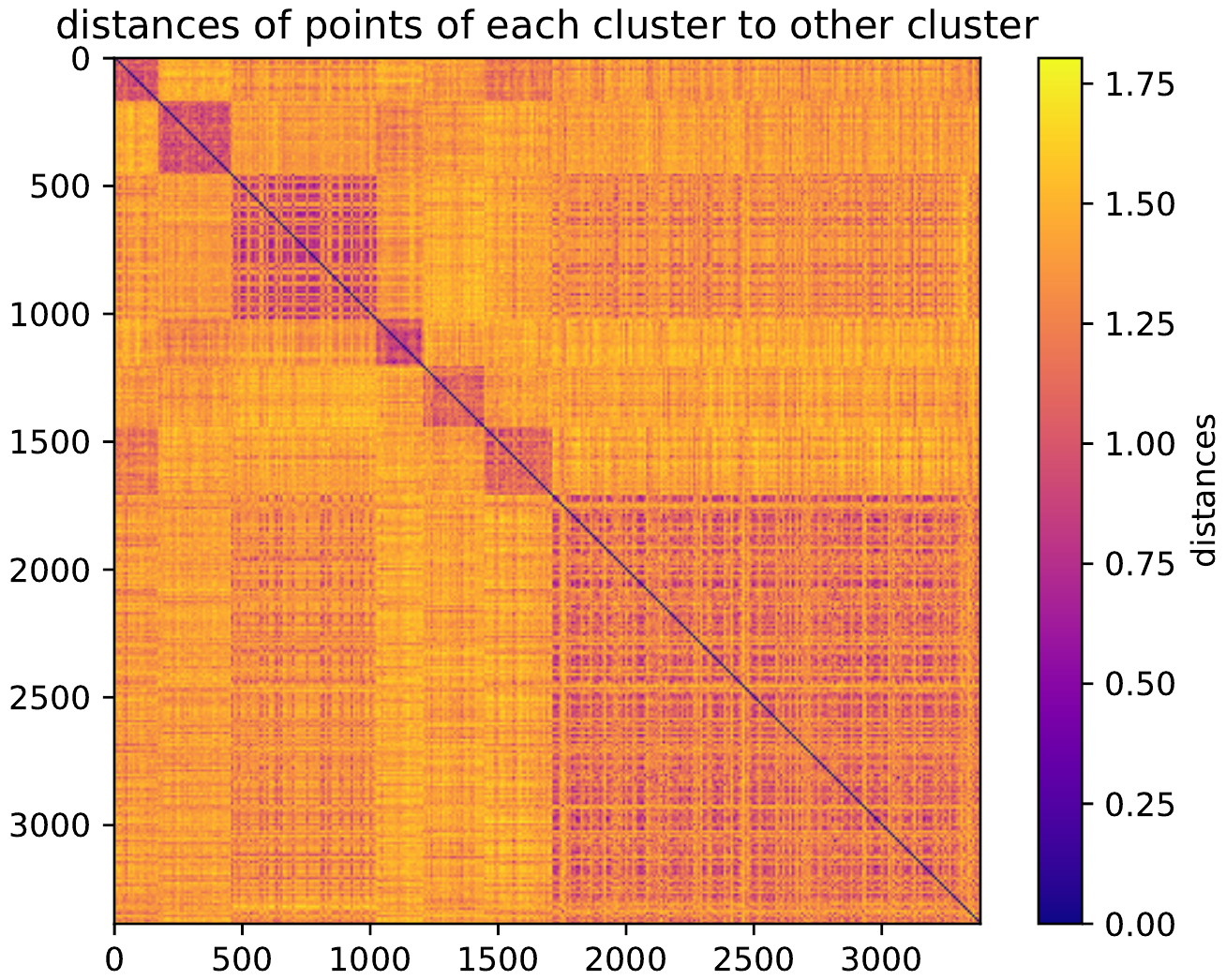}
  \caption{Contextual}
\end{subfigure}
\caption{The distance of each face embedding in \textsc{SC-Friends} to all other face embeddings sorted by cluster after the second stage with tight and contextual bounding boxes. Larger face regions with contextual information results in lower intra-cluster distances and larger inter-cluster distances.}
\label{fig:bb_size}
\vspace{-3mm}
\end{figure}

Additionally, we calculated the Euclidean distance between all pairs of face embeddings, for tight and contextual boxes. The distances are sorted based on the unique named entity assigned to each face after the second stage. As shown in Figure \ref{fig:bb_size}, clearer clusters are formed in the contextual box setting, showing exactly the $7$ character entities in \textsc{SC-Friends}, while this is not the case for tight boxes; the intra-cluster distances are lower (more purple color) and the inter-cluster distances are higher (more yellow).

\subsubsection{Number of clusters} 
\begin{figure}
    \centering
    \includegraphics[width=0.6\linewidth]{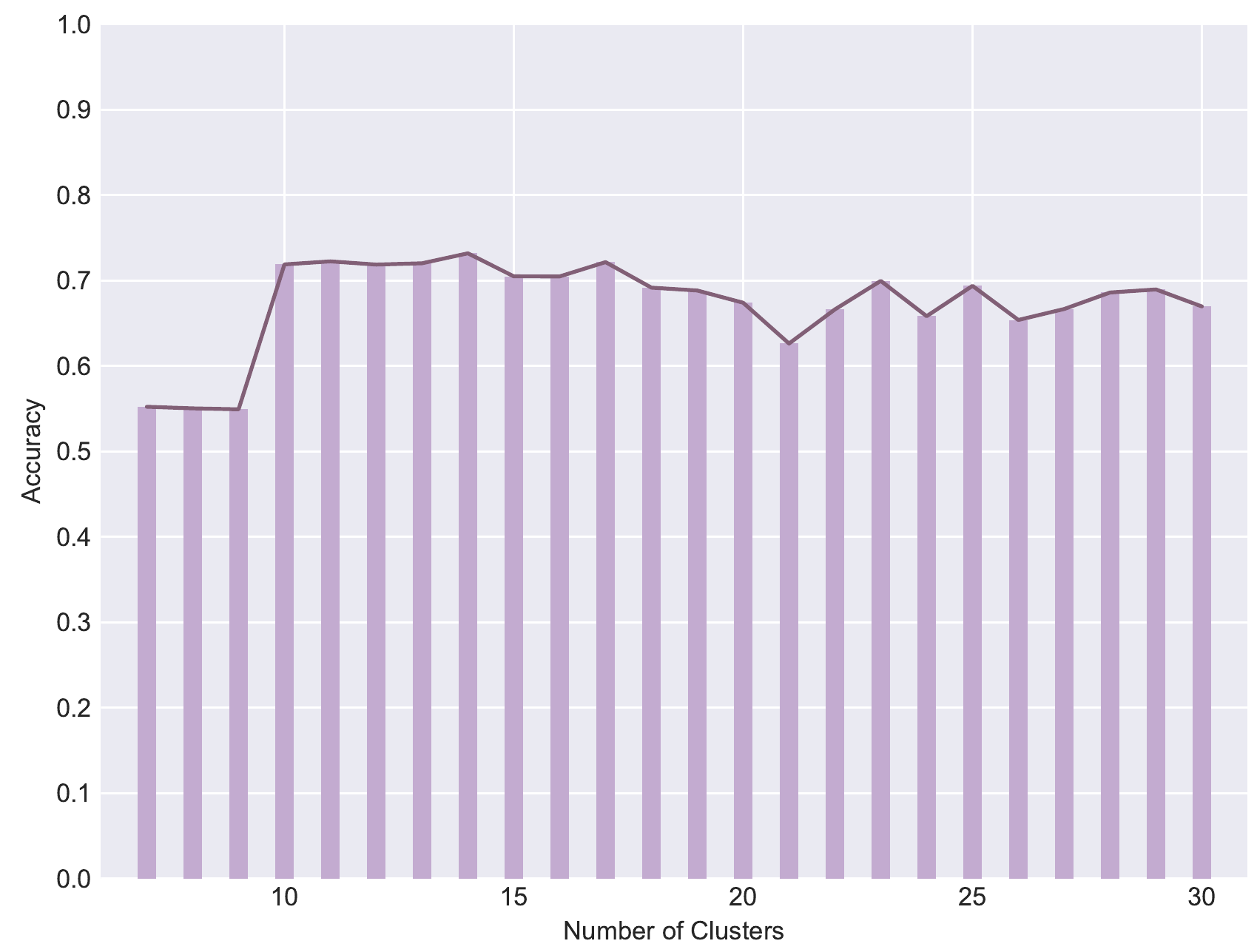}
    \caption{Impact of the number of clusters in \textsc{SC-Friends}. We find that our approach is robust to the number of clusters, as long as the number of clusters is bigger than the number of entities considered in the evaluation.}
    \label{fig:number_of_clusters}
\end{figure}

Here, we investigated the influence of the number of clusters in the second stage of our method. In Stage 2, the face embeddings are over-clustered and then the aggregate bipartite labels from Stage $1$ are assigned for each cluster. Since in the \textsc{SC-Friends} dataset there are $7$ entities, we carry out the Agglomerative clustering in stage $2$ with different numbers of clusters. Figure \ref{fig:number_of_clusters} shows the macro accuracy, on $7$ to $30$ number of clusters. As shown, while a low number of clusters (from $7$ to $10$) reach an obvious low accuracy, the result for a higher number of clusters stays almost the same. Thus, the method appears robust to cluster count as long as there are more clusters than the number of main entities.

\subsection{Qualitative Results}

\begin{figure*}
    \centering
    \includegraphics[width=\linewidth]{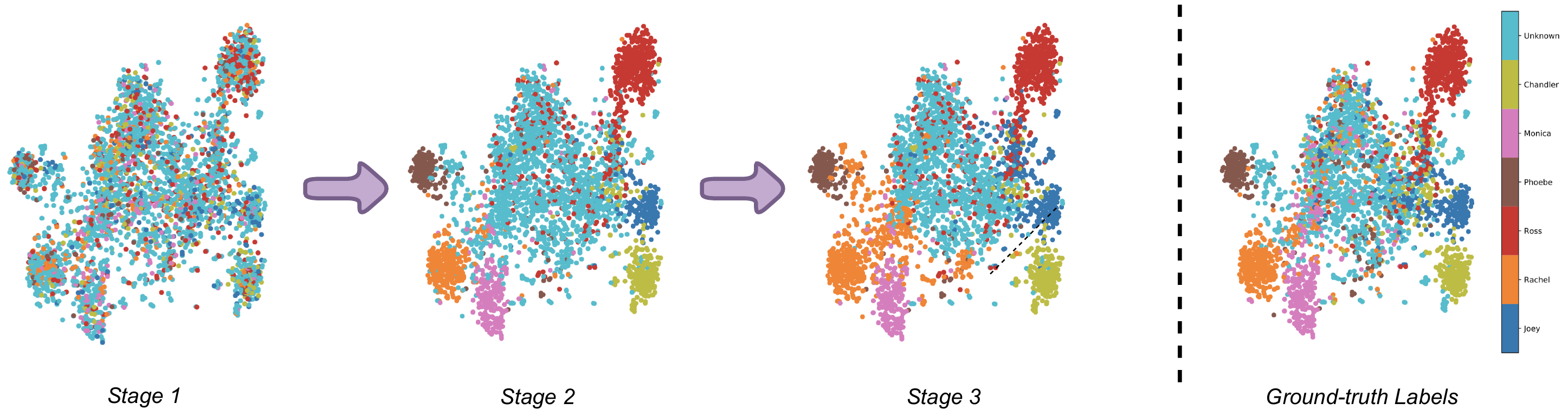}
    \caption{TSNE plots of face embeddings with their predicted labels after each stage of our method for the \textsc{SC-Friends} dataset.}
    \label{fig:tsne}
\end{figure*}

\begin{figure}
    \centering
    \includegraphics[width=\linewidth]{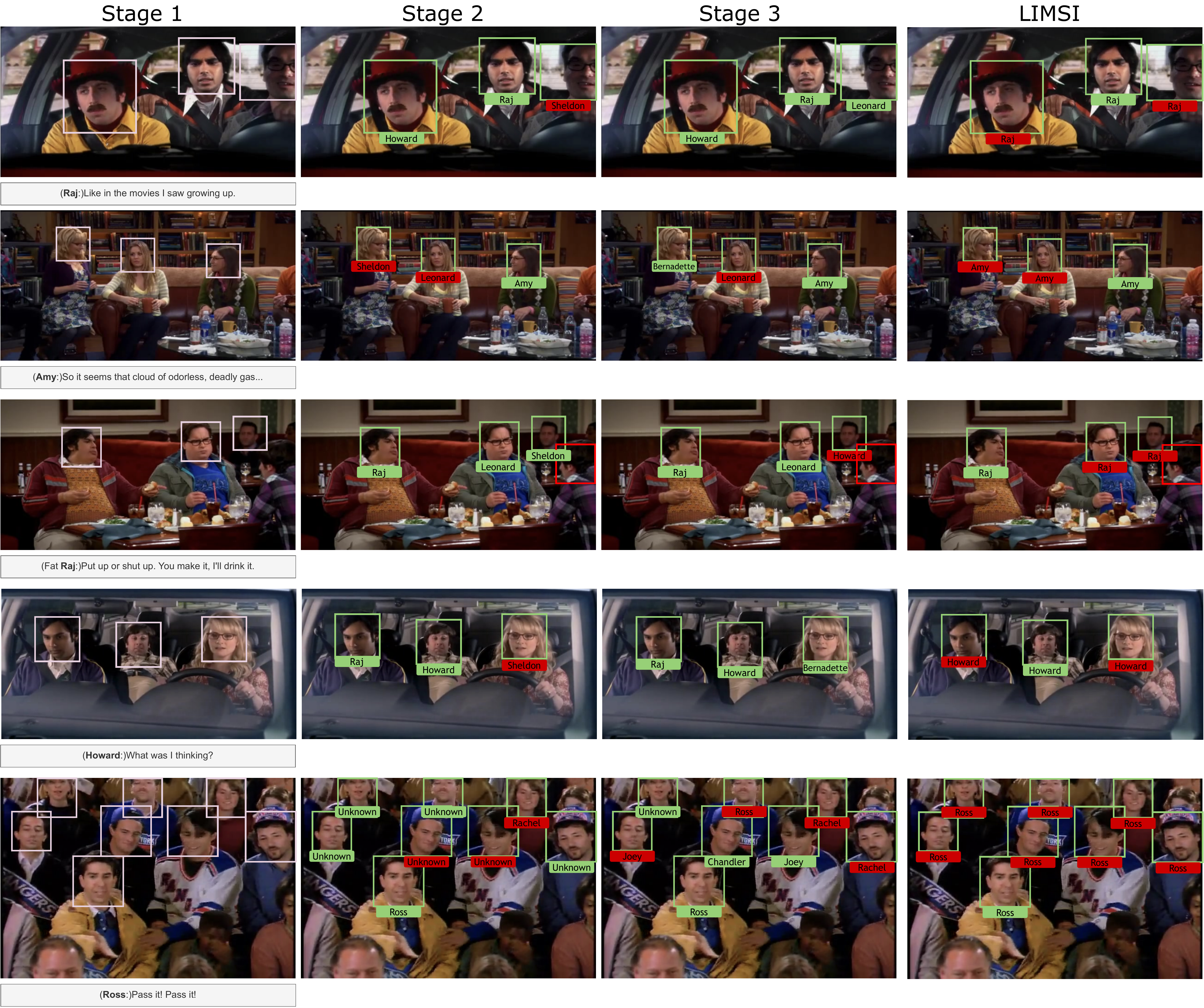}
    \caption{Example predictions for Stage 1, 2, 3 and the LIMSI baseline (Columns 1, 2, 3 and 4 respectively). Green denotes correct and red incorrect. The stage 1 detected named entities and faces are bolded on the subtitles and video frames respectively.}
    \label{fig:qualitative_2}
\end{figure} 

To further understand the performance of our method, we perform a number of qualitative analyses. 
First, we examined the impact of each stage in the face embedding space. In Figure \ref{fig:tsne}, we provide TSNE plots of the embedding space after each stage of method together with a plot of the ground truth embeddings for the \textsc{SC-Friends} dataset.  
From left to right, each plot shows the predicted label after stages 1, 2, and 3 respectively. As is evident, the bipartite labels from the first stage are noisy and it looks like there is little pattern to the predictions, however, some slight signs of clusters for entities such as Ross, Phoebe, and Chandler can be found. The second stage separates entities into clear clusters. The large imbalance between the \emph{unknown} (light blue) and other entities is evident and by comparing predictions to ground-truth labels, we can observe that still many faces that must be predicted as other clusters are predicted as \emph{unknown}. The third stage prediction depicts how it trades-off precision for recall. Samples on the \emph{unknown} boundary are switched to another class, thus, achieving higher class accuracies for entities other than \emph{unknown}. This movement away from the \emph{unknown} class is also evidenced in Table \ref{tab:friends_per_class}. Indeed, these plots again emphasize the importance.
     
\begin{figure}
    \centering
    \includegraphics[width=\linewidth]{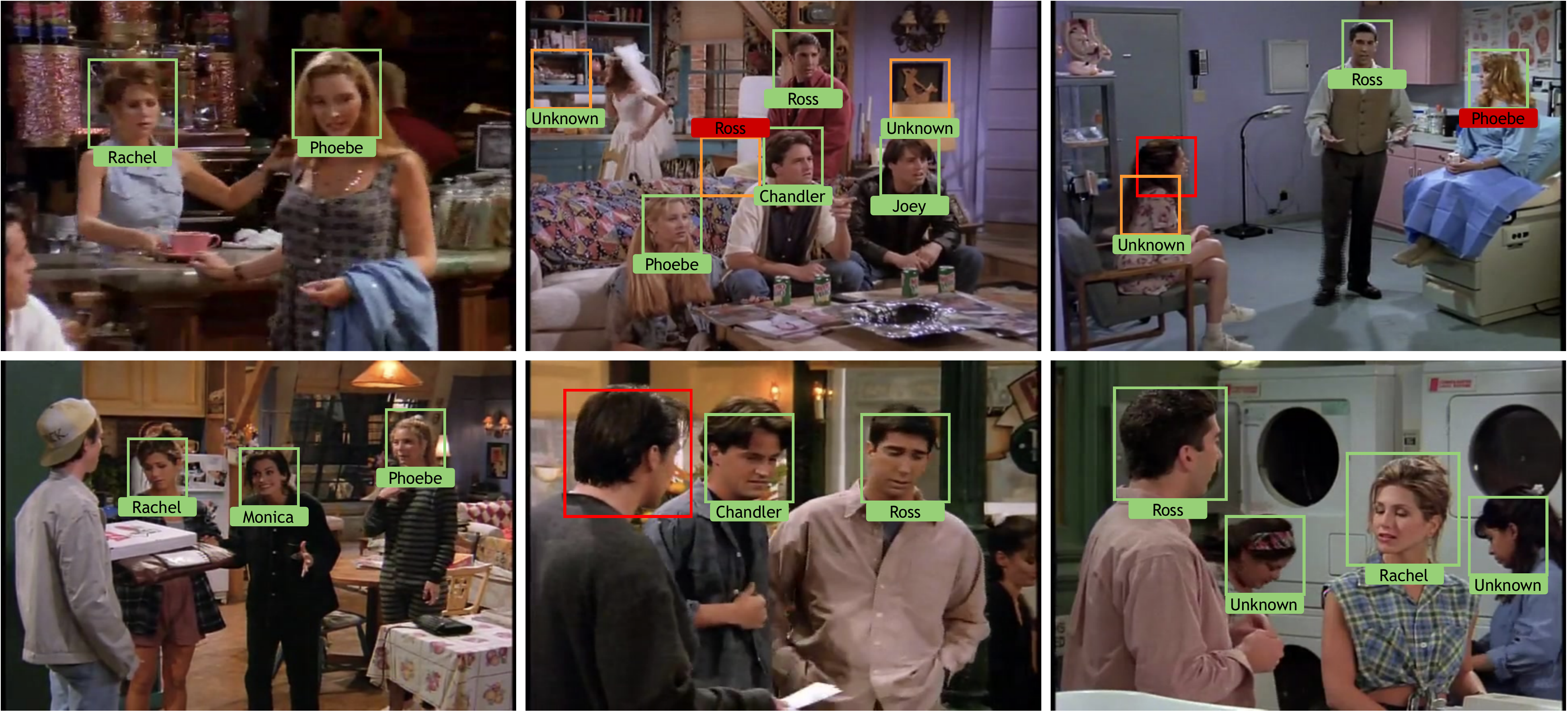}
    \caption{Final predictions from different frames, highlighting both success and failure cases.}
    \label{fig:qualitative_4}
\end{figure}

\begin{figure}
    \centering
    \includegraphics[width=0.7\linewidth]{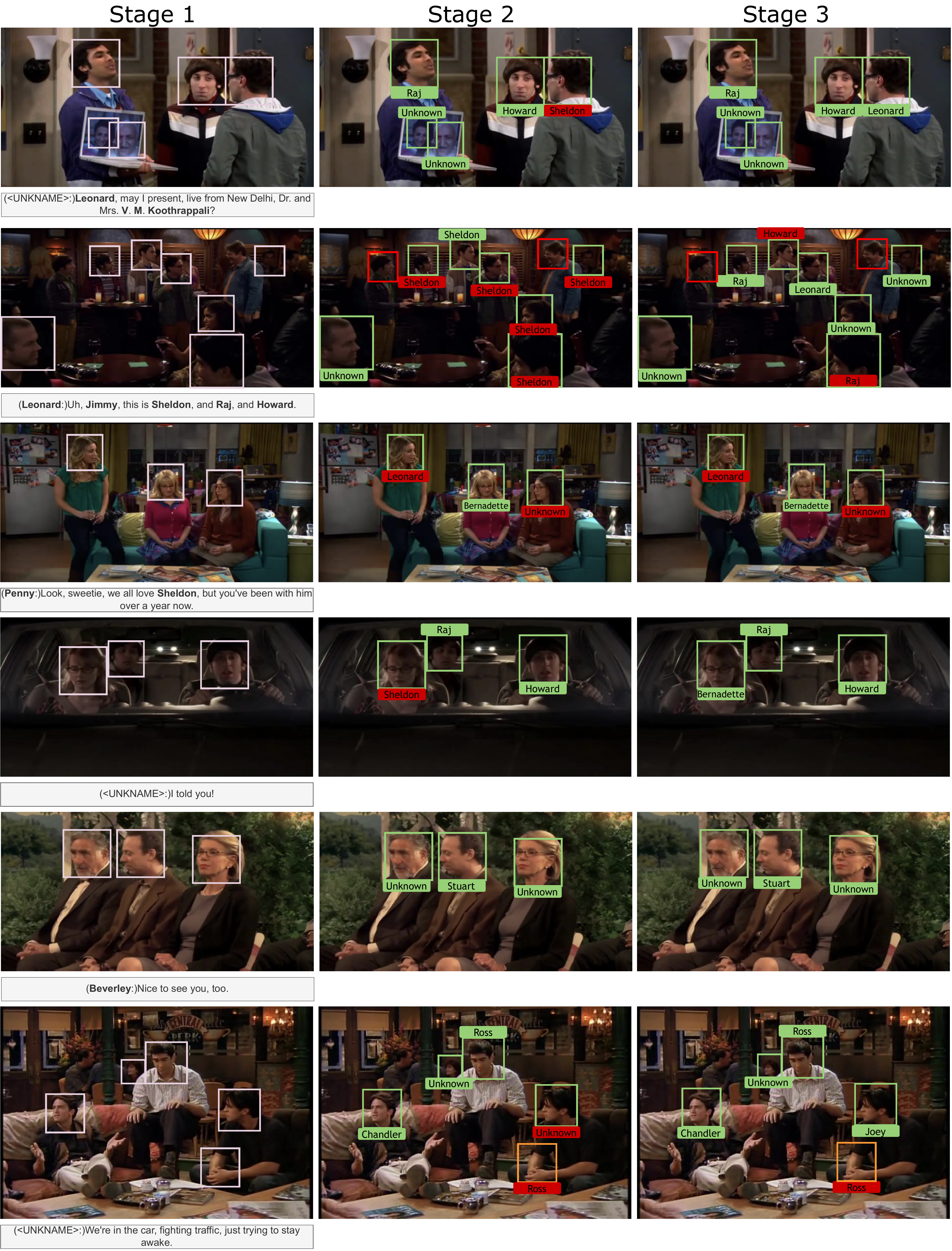}
    \caption{Additional example predictions for each stage of our approach.}
    \label{fig:qualitative_rest}
\end{figure}

Second, we analyzed a number of success and failure cases. In Figure \ref{fig:qualitative_2}, the first column shows the original frame and its corresponding subtitle plus the detected face boxes and the detected entities (bold in subtitle) from our method. The second and third columns show the second and third stage predictions respectively. The fourth column shows the predictions from the LIMSI baseline that in the second stage, matches faces directly to the named entity, if only one name is detected and otherwise runs the same as the second stage of our method. Green box depicts a correct face box and red, a missing face box. The labels per box show the prediction of that stage and the color marks correct or incorrect predictions. Figure \ref{fig:qualitative_4}, shows final example predictions from multiple frames in the series. Orange box shows a non-face bounding box. You can find extra success and failure qualitative results in Figure \ref{fig:qualitative_rest}.

These qualitative results exemplify how our method takes advantage of the multimodal nature of video and focus not just on the video or the caption. The name of most correctly predicted face entities is not present in the subtitle. Hence, taking advantage of the entire videos content. Even in crowded scenes with extras in the background, the main characters are detected and labeled correctly (\eg Fig. \ref{fig:qualitative_4}, second column; top). It is interesting to see that even in different illumination conditions, variations of the same person can be detected correctly (\eg Fig. \ref{fig:qualitative_2}, third row and Fig. \ref{fig:qualitative_rest}, fourth row). 

Turning to failure cases, we see that the lack of a constraint that faces in the same frame need to have different names can lead to incorrect results (\eg predicting Ross twice in Fig. \ref{fig:qualitative_4}, second column; top). This should be taken into account in future work. Another problem, already mentioned in Section \ref{sec:evaluation_of_different_stages}, is the entity Penny is hard for the named entity recognition model to detect, thus it is misclassified in the second and third stage. This is seen in the second and third row of Fig. \ref{fig:qualitative_2} and Fig. \ref{fig:qualitative_rest} respectively.  Additionally, because Stage 3 optimizes for the recall of entities other than \textit{unknown}, we can observe it modifies correctly classified labels from Stage 2. This is seen in  Fig. \ref{fig:qualitative_2}, Row 3 where it incorrectly relabels \emph{unknown} to Howard. However, this can also improve performance as we see in Fig. \ref{fig:qualitative_2}, Row 2 where \emph{unknown} is relabeled to Amy. We provide more qualitative examples in the supplementary material.

Overall, the results demonstrate the strong performance of our method but also the challenging nature of the task. Avenues for future improvement include taking even more advantage of regularities in the video as well as leveraging latent relational information.

\subsection{Qualitative Scene Entity Discovery}
\begin{figure}
    \centering
    \includegraphics[width=\linewidth]{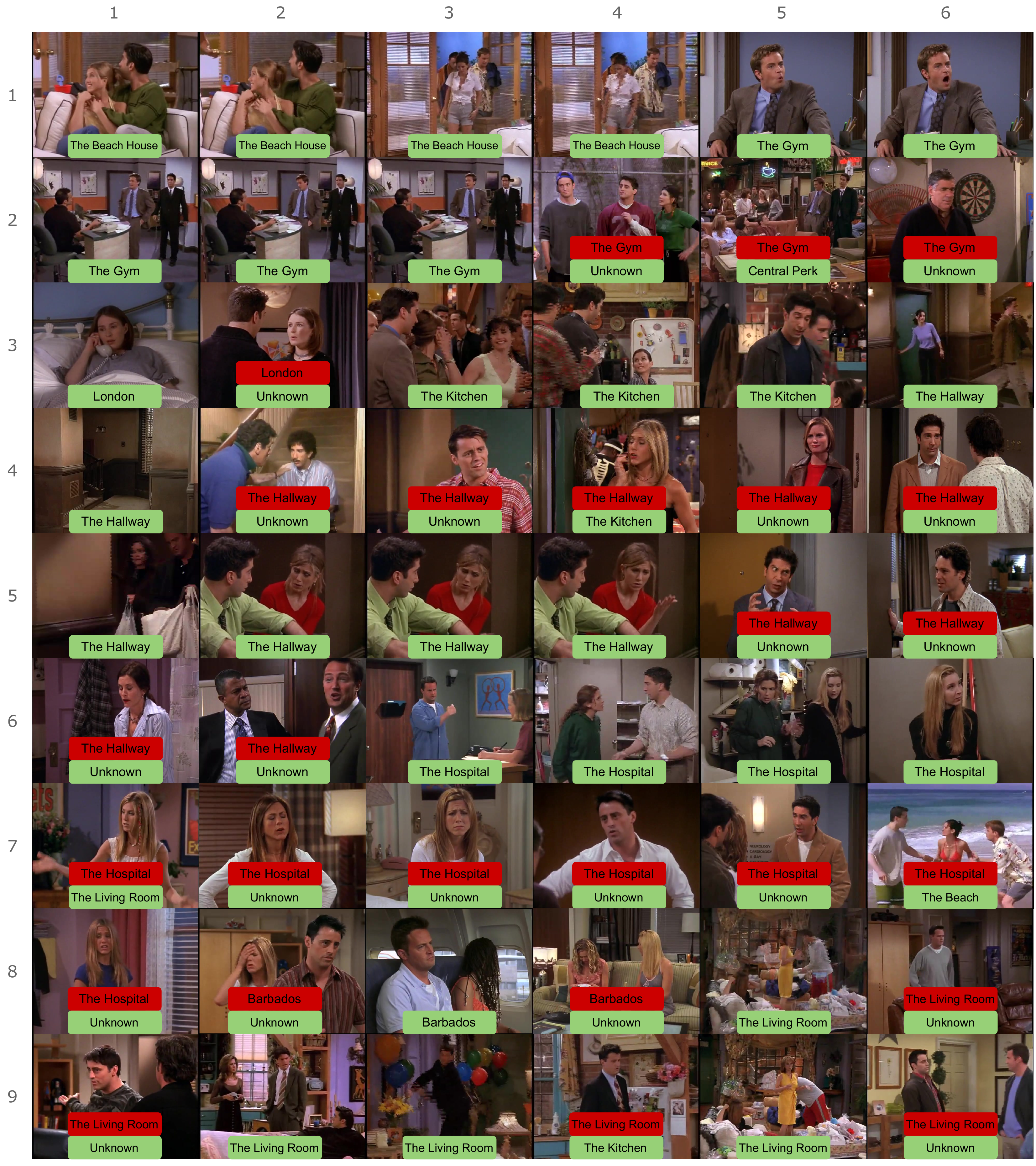}
    \caption{Example scene entity predictions from the \textsc{SC-Friends} dataset. Correct predictions and ground-truth entities are shown in green labels and incorrect predictions in red labels.}
    \label{fig:scenes}
\end{figure}

Thus far, all our experiments focused on faces as visual entities and character names as their textual named entities. In this experiment we demonstrate that our method is generalizable to other entity types. Here, we analyze qualitative success and failure cases with scene visual entities. Figure \ref{fig:scenes} showcases the correctly predicted scene entity in green label for each frame. If the prediction is wrong, both the prediction and the groundtruth scene entity are shown in red and green labels respectively. 

Looking at success cases, we see that the model can discover rare scene entities such as the beach house (from Season 3, episode 25) in row 1, columns 3 to 4 r. 1, c. 3-4 or the gym (from Season 4, episode 4) in r. 1, c. 5-6. It can also detect more common visual entities such as the hallway in r. 4\&5, c. 1. Other success cases include frames with no visual clues and correct predictions such as the well-known hallway in r. 5, c. 2-4 where we only see the characters and the background does not give unique visual information, the hospital in r. 6, c. 3-6, and the gym in r. 2, c. 1-3 with a simple blue background. This predictions are only possible because of the correspondence of the frames and mentions in subtitles. In some examples minimal visual cues are available such as the living room in r. 9,c. 3, Monica's kitchen in r. 3, c. 3 and r. 3, c. 5, London in r. 3, c. 1, the beach house in r. 1, c. 1-2, or Barbados in r. 8, c. 3.

On the other hand, by looking at failure cases, we see that some visual clues possibly distract the model prediction such as the dart board in r. 2, c. 6, or the doors and steps   that visually suggest the hallway prediction r. 4 c. 2-6 and r. 5, c. 5-6. The frame in r. 2, c. 4 are predicted as gym  because of their cloths and that they are playing football. Frames in r. 7, c. 1-5 are biased towards the hospital prediction probably because of wearing white cloths that look like doctor uniforms. In other failure cases we observe textual cues that distract the model like in r. 3, c. 2 where the characters talk about moving to London, or in r. 7, c. 6 where there is an obvious beach but because of the Jelly fish sting the characters talk about the hospital. The same holds for Barbados prediction in  r. 8, c. 2\&4 and r. 2, c. 5 that the visual cues are very strong for predicting the coffee house Central Perk. Another source of failure is the pretrained generic visual models that are trained to distinguish beach from living room and not distinguish between different living rooms such as scene instance detection. These models collapse the embeddings of different e.g. living rooms close to each other, thus not detecting multiple instances of living room such as in r. 8, c. 6 and r. 9,c. 1 that is Chandler and Joey's living room and not Monica's, or in r. 9, c. 6 Phoebe's living room. Moreover, the pretrained models on places are not robust enough to work for real scenarios, and close-up camera poses with multiple people in the frame. Another challenge is filtering the unique scene entity mentions from all mentions. For instance some mentions refer to someone's house by saying my house or your house but need extra co-reference to resolve both of the mentions into e.g. Monica's place.

Overall, this experiment shows the potential and the challenges of the video entity discovery task and how it can be applied to different entity types. The challenge can be solved from different perspectives such as improving the mention detectors, visual detectors, instance recognition and inspires more research in the field.

\section{Conclusion}
This work investigates how to discover named entities in videos in a self-contained manner. Rather than relying on entity-specifc annotations or knowledge sources, we show that the multimodal nature of captioned videos on its own already provides a fruitful source to learn which entities belong to which visual entity (e.g. face). To that end, we introduce a three-stage method, where we first obtain visual entities and named entities separately from video frames and captions, followed by a bipartite alignment between entities and visual entities occurring at the same time. This matching provides an initial clue but is noisy and incomplete. As a second stage, we over-cluster visual entities in their embedding space and use the bipartite matching to aggregate the most prominent entity per cluster. As a final balancing stage, we construct prototypes in the visual embedding space and reassign entities based on their distances to the prototypes. We performed empirical evaluations on two new benchmarks, \textsc{SC-Friends} and \textsc{SC-BBT}. \textsc{SC-Friends} is manually curated, while \textsc{SC-BBT} is a larger-scale benchmark with annotations derived from a video question and answering task. On both datasets, we find that our approach is capable of discovering entities with results competitive to supervised baselines. These results and our qualitative analysis highlight that it is indeed possible to discover entities in a self-contained manner without entity-specific human intervention. For future work, we are keen to investigate self-contained entity discovery in videos beyond TV series such as news items.







\bibliographystyle{ACM-Reference-Format}
\bibliography{sample-base}




            

\end{document}